\documentclass[10pt, conference]{IEEEtran}
\IEEEoverridecommandlockouts
\usepackage{cite}
\usepackage{amsmath,amssymb,amsfonts}
\usepackage{algorithmic}
\usepackage{graphicx}
\usepackage{textcomp}

\usepackage{algorithm}
\usepackage[left=0.68in, right=0.673in, top=0.78in, bottom=1.1in]{geometry}

\usepackage{xcolor}
\def\BibTeX{{\rm B\kern-.05em{\sc i\kern-.025em b}\kern-.08em
    T\kern-.1667em\lower.7ex\hbox{E}\kern-.125emX}}
\begin{document}

\title{Fault Detection in Mobile Networks Using Diffusion Models}

\author{\IEEEauthorblockN{Mohamad Nabeel\,, 
Doumitrou Daniil Nimara\,, 
Tahar Zanouda }\
\IEEEauthorblockA{Global AI Accelerator, Ericsson, Sweden}\
\texttt{\{mohamad.nabeel, doumitrou.nimara, tahar.zanouda\}@ericsson.com}}

\maketitle

\begin{abstract}

In today's hyper-connected world, ensuring the reliability of telecom networks becomes increasingly crucial. Telecom networks encompass numerous underlying and intertwined software and hardware components, each providing different functionalities. To ensure the stability of telecom networks, telecom software, and hardware vendors developed several methods to detect any aberrant behavior in telecom networks and enable instant feedback and alerts. These approaches, although powerful, struggle to generalize due to the unsteady nature of the software-intensive embedded system and the complexity and diversity of multi-standard mobile networks.

In this paper, we present a system to detect anomalies in telecom networks using a generative AI model. We evaluate several strategies using diffusion models to train the model for anomaly detection using multivariate time-series data. 
The contributions of this paper are threefold: (i) A proposal of a framework for utilizing diffusion models for time-series anomaly detection in telecom networks, (ii) A proposal of a particular Diffusion model architecture that outperforms other state-of-the-art techniques, (iii) Experiments on a real-world dataset to demonstrate that our model effectively provides explainable results, exposing some of its limitations and suggesting future research avenues to enhance its capabilities further.

\end{abstract}


\section{Introduction}
\label{section:intro}
    The telecom sector recognizes the critical importance of ensuring the stability, resilience, and reliability of Radio Access Networks (RAN) before deployment in operational environments. Telecom software (SW) vendors strive to provide efficient
    and robust SW to ensure a seamless user experience. However, faults in the SW can occur for a variety of reasons. To assess the stability of the SW for the operational network, extensive testing is performed in a controlled environment. While the SW may show minimal or no erroneous behavior during these evaluation tests in a controlled environment, silent anomalous behavior can still exist for any reason after deployment. Telecom SW vendors are actively looking for ways to improve fault detection methods to increase the efficiency of products. Generative models \cite{chang2023design} have been widely used in the domain for a while, and Denoising Diffusion Probabilistic Models (DDPM), also called diffusion models \cite{yang2023diffusion}, have risen in popularity, often outperforming the existing generative alternatives. Initially introduced by Dickstein et al., \cite{diffusion2015}, diffusion models gained further prominence when Ho et al., \cite{denoisingdiffusion} presented high-quality image synthesis results using these models. As a result, diffusion models have been widely applied in various data domains. However, as of yet, even though there are plenty of promising DDPMs that handle time-series (TS) data well \cite{lin2023diffusion}, there hasn't been any research done on the effectiveness of applying DDPM for telecom network fault detection tasks. Therefore, the primary objective of this paper is to advocate for the efficacy of DDPM in TS and RAN data, exploiting their potential as fault detectors. By leveraging the capabilities of DDPMs, the aim is to accelerate fault detection, reduce response times, and efficiently resolve problems in the software deployment process. Integrating AI techniques, particularly diffusion models, is expected to improve fault detection in the telecom industry \cite{bariah2024large}, delivering more reliable and seamless services to end users.
    

    The contribution of this paper is threefold. We present:
    \begin{enumerate}
        \item A framework to identify anomalies in Telecom networks using diffusion models.
        \item A suitable Diffusion model Architecture that is competitive with alternative solutions.
        \item Experiments and results, accompanied by analysis, that provide insights on model performance, exposing some limitations and suggesting future enhancement plans.
    \end{enumerate}

    Overall, this work pioneers the usage of Diffusion models Fault Detection in Telecom Networks and paves the way for a fruitful and promising research field.
\section{Related work}
    Anomaly Detection \cite{GNN} refers to identifying anomalous behavior where observations deviate significantly from the normal behavior. In this section, we explore the utilization of deep learning models for (i) Forecasting-based and (ii) Reconstruction-based anomaly detection.
            \subsection{Forecasting-based anomaly detection}
            In Forecasting-based anomaly detection, the generated output is conditioned on historical data. It is expected to reconstruct the remaining masked part of the input sample. Different approaches have been developed.  
                Ergen et al. \cite{LSTM-based} focused on unsupervised anomaly detection and proposed algorithms utilizing Long Short-Term Memory (LSTM) neural networks. They utilize variable-length data sequences through an LSTM-based structure to obtain fixed-length sequences. The anomaly detection is based on decision functions derived from One-Class Support Vector Machines (OC-SVM) and Support Vector Data Description (SVDD) algorithms \cite{LSTM-based}. Ergen et al. \cite{LSTM-based} modify the original objective criteria of OC-SVM and SVDD algorithms to enable gradient-based training. The proposed algorithms yield notable improvements.

                Ren et al. \cite{SRCNN} proposed an anomaly detection service developed by Microsoft, focusing on real-time monitoring of various metrics (e.g., Page Views and Revenue) for applications and services. The service employs a TS anomaly detection approach to monitor and promptly alert users about potential incidents continuously. The proposed pipeline encompasses three main modules: data ingestion, an experimentation platform, and online computation, aiming for accuracy, efficiency, and generality in design. They introduced a novel algorithm for TS anomaly detection, leveraging Spectral Residual (SR) and Convolutional Neural Network (CNN) techniques \cite{SRCNN}. According to Ren et al. \cite{SRCNN} this approach is the first application of the SR model, initially used in visual saliency detection, to TS anomaly detection. The combination with CNN enhanced the SR model's performance \cite{SRCNN}.
                
                Bourgeri et al. \cite{GNN} addressed the challenge of detecting faults in telecom networks by leveraging the intertwined nature of embedded and software-intensive Radio Access Network systems, using combined telecom network topology and SW execution graphs. The proposed solution involves utilizing bi-level Federated Graph Neural Networks to identify anomalies in the telecom network. The model was evaluated under three settings:
                (i) Centralized Temporal Graph Neural Networks Model: This model detects anomalies in 4G/5G telecom data in a centralized training setting \cite{GNN}.
                (ii) Federated Temporal Graph Neural Networks Model: Which utilizes Federated Learning
                (iii) Personalized Federated Temporal Graph Neural Networks Model: A novel aggregation technique, referred to as FedGraph, is proposed, leveraging both graphs and site similarities for aggregating models and tailoring them to each site’s behavior \cite{GNN}.
                
            \subsection{Reconstruction-based Anomaly Detection}
            
        In Reconstruction-based anomaly detection, the data is fully masked, and the model is expected to generate a sample without any conditional information presented. 
                González et al. \cite{netgan} introduced Net-GAN and Net-VAE for network anomaly detection in multivariate TS data. To capture the temporal correlations characterized in multi-variate TS cases, the original GAN model's multilayer perceptrons are replaced by recursive LSTM networks for both the generator and discriminator model \cite{netgan}. 
                They also explore the performance of using VAE; in their paper, they call it Net-VAE \cite{netgan}. They evaluate these two models in different monitoring scenarios, including anomaly detection in Internet of Things sensor data and intrusion detection in network measurements \cite{netgan}.

                Garcia et al. \cite{tstoimg} stated that essential characteristics of time-series situated outside the time domain are often difficult to capture with the SOTA anomaly detection without transformations on TS data. Promising results have been achieved by transforming time series into image-like representations. The paper first reviews different signal-to-image encoding approaches in the literature. Secondly, they propose modifications to make the encodings more robust to variability in larger datasets. Third, they demonstrate how the choice of encoding can impact the result using the same deep learning architecture \cite{tstoimg}. Six encoding algorithms were selected: Gramian Angular Field, Markov Transition Field, recurrence plot, greyscale encoding, spectrogram, and scalogram \cite{tstoimg}. They showed that these encodings have a competitive advantage and might be worth using in the DL framework \cite{tstoimg}.
                In Telecom context, Zanouda et al. \cite{FaultDetectionLSTM} used LSTM Autoencoders to identify and locate anomalies in mobile networks using software performance TS data. 
                Recently, early applications of Diffusion-based models  \cite{pintilie2023time} \cite{2305.18593} \cite{2310.08800} on TS \cite{lin2023diffusion} data showed promising results. 



\section{Data}
\label{section:data}

A telecom operational network consists of interconnected Radio nodes that provide network coverage, where each Radio node (e.g., gNB, eNB, etc) is supported by a SW to handle various network functionalities. Each component in the network is monitored to ensure network stability.

The data we use in this paper is a private dataset of internal SW checkpoints, accessible only to Telecom SW vendors, collected as part of the internal logging system of a Telecom SW Vendor (Ericsson). The data was collected from 3 representative areas, (1) a downtown area, (2) an airport area, and (3) a suburban area to assess SW quality. The data was collected from 67 RAN nodes, of which 12 of them were from the airport area, 29 of them were from the downtown area, and 26 of them were from the suburban areas. Because of the confidential nature of the data, it can only be stated that the data is TS data collected from different SW NR/LTE procedures of the same SW deployed on multiple Radio Nodes. The dataset comprises 263,700 samples with 17 features (SW checkpoints) where $1\%$ were annotated as silent anomalies.

\section{Methods}

\subsection{Problem Formulation}

\noindent\textbf{Telecom Network Radio Nodes:}
The network consists of $N$ telecom RAN nodes evolving overtime:
\begin{equation}
    R=\{R_1, \ldots, R_N\}
\end{equation}
 where $R$ represents the set of RAN nodes in an operational telecom network $NW_{operator}$.

\noindent\textbf{Multivariate TS:}
For each RAN node $R_{i}$ in telecom network $NW_{operator}$, we consider a multivariate time-series of $K$ features with the TS-length of $L$:
\[(\mathbf{x}^l_0)_{l=1}^{L} = (x_1^l, \ldots, x_K^l)_{l=1}^{L}\]
where each data point $x_k^l\in \mathbb{R}$ represent a value that is collected at timestamp $l$ for the SW feature $k$.

\noindent\textbf{Anomaly Detection:}
Given a temporal input with a time length of $L$, we aim to predict:
\[(\mathbf{y}^l)_{l=1}^{L} = \{y_1^l, \ldots, y_K^l\}_{l=1}^{L}\]
where $y_k^l \in \{0, 1\}$ denotes whether the data point at the timestamp of the node $x_k^l$ is anomalous ($1$ denotes an anomalous data point).

\subsection{Overview of the process}
%
%
%
%
%

\begin{figure}[h]
            \centering
            \includegraphics[scale=0.25]{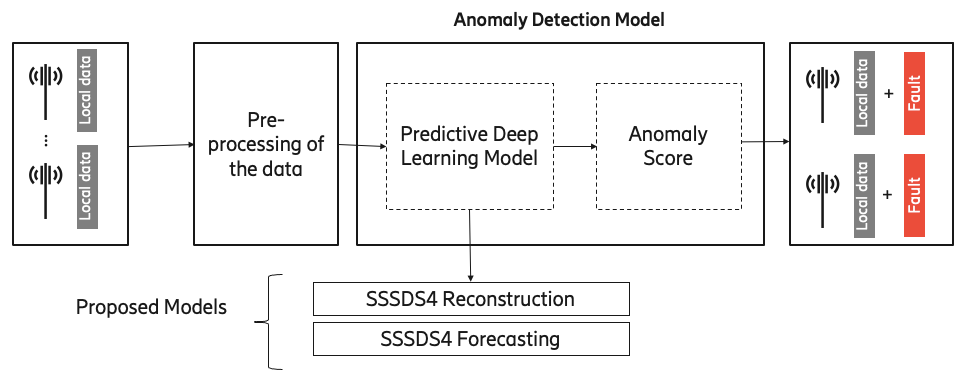}
            \caption{The process of how the data is used for anomaly detection.}
            \label{fig:forecasting_reconstruction}
        \end{figure}

         The pipeline consists of three main steps as depicted in Fig. \ref{fig:forecasting_reconstruction}. We propose two types of architectures for anomaly detection, namely, (i): reconstruction-based and (ii) forecasting-based. The model was trained in an unsupervised manner. The labeled samples from the dataset were used in the evaluation phase to quantify the performance of the anomaly detector.

         \begin{algorithm}
                \caption{Training}
                \label{alg:algo_diffusion_1}
                \begin{algorithmic}[1]
                    \REPEAT
                    \STATE{$\mathbf{x}_{0}^{l} \sim q(\mathbf{x}_{0}^{l})$}
                    \STATE{$t \sim$ Uniform$(\{1,\ldots,T\})$}
                    \STATE{$\boldsymbol{\epsilon} \sim \mathcal{N}(\mathbf{0},\mathbf{I})$}
                    \STATE{Take gradient descent step on \\
                        $\quad \nabla_\theta||\boldsymbol{\epsilon}-\boldsymbol{\epsilon}_{\theta}(\sqrt{\bar{\alpha}_{t}}\mathbf{x}_{0}^{l} + \sqrt{1 - \bar{\alpha}_t}\boldsymbol{\epsilon}, t)||^2$}
                    \UNTIL{converged}
                \end{algorithmic}
                    
            \end{algorithm}
         
         \subsubsection{Data pre-processing}
    
            In the first step of the pipeline, during the pre-processing of the data, the data is processed according to the required input format of the network architecture. In the reconstruction-based anomaly detection model, the data format was in $\mathbf{x} \in \mathbb{R}^{B\times C \times L}$ where $B = 2637$ is the batch size, $C = 17$ which was the number of features, and $L = 100$ was the window size.
            For the forecasting-based anomaly detector, the data was sampled by a \textit{sliding window} rather than divided into window sizes. The window size used in this approach was $L = 250$.  

         \subsubsection{Anomaly Detection Model}
                Using the pre-processed data $\mathbf{x}_{0}^{l}$, the model is first trained to generate $\hat{\mathbf{x}}_{0}^{l}$ according to the training algorithm illustrated by Ho et al. \cite{denoisingdiffusion} depicted in Algorithm \ref{alg:algo_diffusion_1}.

 \begin{algorithm}
  \caption{Anomaly Detection}
                    \label{alg:algo_diffusion_2}
                \begin{algorithmic}[1]
                    \STATE{\textbf{Input}: $\mathbf{x}_{0}^{l}$, \textbf{ Threshold}}, 
                    \STATE{$\boldsymbol{\alpha}$: Fixed linear schedule, $\bar{\boldsymbol{\alpha}_t} = \prod_{s=1}^{t}\alpha_s$}
                    \STATE{{\textbf{Output}: F1-score, Precision-score, Recall-score}} \\
                    $// \text{ Use forward pass for Reconstruction}$
                    \IF{reconstruction} 
    
                    \STATE{$\boldsymbol{\epsilon} \sim \mathcal{N}(\mathbf{0}, \mathbf{I})$}
                    \STATE{$\hat{\mathbf{x}}_{T}^{l} = \sqrt{\bar{\alpha}_T}\mathbf{x}_{0}^{l} + \sqrt{1 - \bar{\alpha}_T}\boldsymbol{\epsilon}$}
                    \STATE{\text{Context}$ = 0$}
                    \ENDIF
    
                    $// \text{ Condition on history of length $h$}$
                    \IF{forecasting}
                    \STATE{$\hat{\mathbf{x}}_{T}^{l} \sim \mathcal{N}(\mathbf{0}, \mathbf{I})$}
                    \STATE{Context = $q(\mathbf{x}_{0}^{[:h]})$}
                    \ENDIF
    
                    \FOR{$t = T, \ldots, 1$}
                        \STATE{$\mathbf{z} \sim \mathcal{N}(\mathbf{0}, \mathbf{I})$ if $t > 1,$ else $\mathbf{z} = \mathbf{0}$}
                        \STATE{$\hat{\mathbf{x}}_{t-1}^{l} = \frac{1}{\sqrt{\alpha}_t}\left(\hat{\mathbf{x}}_{t}^{l} - \frac{1-\alpha_{t}}{\sqrt{1 - \Bar{\alpha}_t}}\boldsymbol{\epsilon}_{\theta}(\hat{\mathbf{x}}_{t}^{l}, t, \text{Context})\right) + \sigma_{t}\mathbf{z}$}
                    \ENDFOR
                    \STATE{$\mathbf{Z} =$ Z-score($\mathbf{x}_{0}^{l}, \hat{\mathbf{x}}_{0}^{l}$)}
                    \STATE{$\hat{\mathbf{y}}^l = \mathbf{Z} > \textbf{Threshold}$}
                    \RETURN{F1($\mathbf{y}^l,\hat{\mathbf{y}}^l$), Precision($\mathbf{y}^l,\hat{\mathbf{y}}^l$), Recall($\mathbf{y}^l,\hat{\mathbf{y}}^l$)}
                    \end{algorithmic}
                   
                \end{algorithm}
     
                This applies to both the reconstruction-based and the forecasting-based model. Then, once training is completed, the model is used for anomaly detection. The steps conducted in anomaly detection are depicted in \ref{alg:algo_diffusion_2}.
        In Algorithm  \ref{alg:algo_diffusion_2}, once  $\hat{\mathbf{x}}_0$ is retrieved, it is then used together with the original data $\mathbf{x}_0$ to compute the \textit{Z-Score}. A \textit{threshold} is given for all features included in the data, and samples are notated as an anomaly or not based on the comparison of the values of the \textit{Z-Score} and the \textit{thresholds}. Finally, the evaluation metrics are computed by comparing the annotated samples with the ground-truth values.

    \subsection{Network Architecture}
\begin{figure}[!ht]
            \centering
            \includegraphics[scale=0.6]{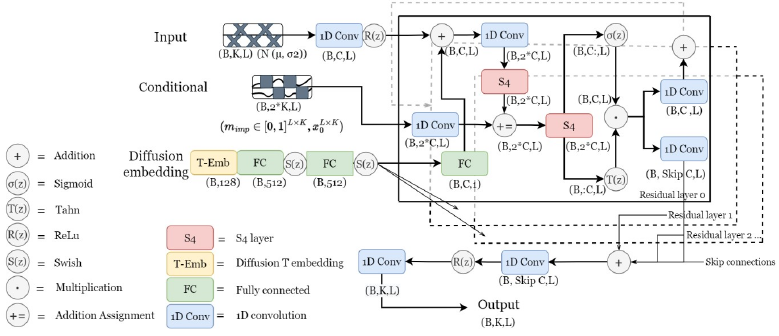}
            \caption{An overview of the SSSDS4 by Miguel et al. \cite{SSSD}.}
            \label{fig:SSSD}
        \end{figure}

        The network architecture for this task is the Structured State Space Sequence Diffusion Model (SSSDS4) \cite{SSSD} proposed by Miguel et al. \cite{SSSD}. An overview of the network architecture can be seen in Figure \ref{fig:SSSD}. 
        This architecture uses Structured State Space Sequence (S4) for modeling the long-range dependencies and sequences that are given. The first S4 layer captures the sequences and dependencies from the given input; the second S4 layer models the long-range dependencies and sequences when the conditional information is added along with the input. The Diffusion T-Embedding defines the diffusion timestep the network architecture is currently denoising, as different diffusion timesteps require different outputs. Alongside the input, conditional information, such as a prior subsequence $h$, can be provisioned, allowing us to condition generation for forecasting and data imputation. This conditional information is omitted during reconstruction-based anomaly detection. In forecasting-based anomaly detection, the model was conditioned on the first 175 timesteps $(h = 175)$ and was tasked to predict the remaining 75 timesteps for each sample. The choice of working with these specific timesteps was based entirely on the same parameter used in the paper of Miguel et al. \cite{SSSD}. The same default hyperparameters given in their repository\footnote{https://github.com/AI4HealthUOL/SSSD} are used to generate the data this paper will work with.

    \subsection{Anomaly Detection}
        We leverage \textit{Z-Score} to compare generated and original samples. The \textit{Z-Score} represents the number of standard deviations
        an individual data point is from the mean of a distribution. The \textit{Z-Score}
        enables the standardization of the values in a dataset, transforming them into
        a common scale. Positive \textit{Z-Score} indicate data points above the mean,
        while negative \textit{Z-Score} indicates data points below the mean. The \textit{Z-Score}
        computation offers a standardized representation of data, which eliminates
        the influences of varying scales or units, allowing fair comparisons across
        different features and datasets. It is essential to remember that the \textit{Z-Score} assumes
        that the data follows a normal distribution, as this assumption may not always
        hold, it may still be useful as a starting point for anomaly detection for
        approximately normally distributed or transformed data. Once the model has generated samples, the magnitude of the difference between the original samples (denoted as $\mathbf{x}_i$) and the generated samples (denoted as $\hat{\mathbf{x}}_i$) for each feature $i$ will be computed
        \begin{equation}
            \mathbf{s}_i = |\mathbf{x}_i - \hat{\mathbf{x}}_i| 
        \end{equation}
        the difference $\mathbf{s}_i$ will then be used to compute the \textit{Z-Score} for each sample
        \begin{equation}
            \mathbf{Z}_{i} = \frac{\mathbf{s}_i-\bar{\mathbf{s}}_i}{\sigma_{i}}
        \end{equation}
        Where $\bar{\mathbf{s}}_i$ is the median of the difference between the original and the generated samples at feature $i$, and $\sigma_i$ is the standard deviation of the difference between the original and the generated samples at feature $i$. A \textit{threshold} for each feature will be computed
        \begin{equation}
            Threshold_{i} = \bar{Z}_i + k_{i}*\Sigma_{i}
        \end{equation}
        where $\bar{Z}_i$ is the mean of the \textit{Z-Score} at feature $i$, $k_i$ is a standard deviation coefficient for feature $i$, and $\Sigma_{i}$ is the standard deviation of the \textit{Z-Score} at feature $i$. An anomaly is assumed to be detected if $|Z_i| > Threshold_i$. This measurement will be computed on a held-out test set. The anomaly labeling for each feature made by the \textit{Z-Score} measurement will be concatenated.
        
    \subsection{Evaluation}
        Since experts manually annotate the samples, the performance of the model can be determined by computing the Precision-, Recall-, and F1-score:
        \begin{equation}
                Precision = \frac{tp}{tp+fp}
        \end{equation}
        \begin{equation}
                Recall = \frac{tp}{tp+fn}
        \end{equation}
        \begin{equation}
                F_1 = \frac{2tp}{2tp+fp+fn}
        \end{equation}
        where $tp$ stands for true positive, $fp$ stands for false positive, and $fn$ stands for false negative. Precision allows for evaluating the detector's accuracy in correctly identifying genuine anomalies, thereby assessing its capability to minimize false positives. A high precision score indicates that the anomaly detector has a low rate of incorrectly labeling normal instances as anomalies. Recall measures the proportion of true anomalies that were correctly detected. A high recall score implies that the detector has a low rate of missing actual anomalies.  F1 scores calculate the harmonic mean of precision and recall, which combines the two measures to provide a balanced assessment of the anomaly detector's performance. Since the anomalies are typically rare compared to normal samples, the F1 score addresses this imbalance by accurately measuring the anomaly detector's performance while minimizing false positives, offering an overall comprehensive performance measurement.

    \subsection{Baselines}
        We compare our proposed method against two baselines that are currently used to detect anomalies on the same dataset \ref{section:data}.
        \noindent\textbf{LSTM-AutoEncoder:} The current ways of detecting the faults in the SW are done by applying the LSTM-AutoEncoder as a reconstruction-based anomaly detector. The data is fed into two LSTM layers before it is encoded, and two LSTM layers are introduced after the data is decoded. The embedding dimension was equal to $128$, and the learning rate was equal to $10^{-3}$.

        \noindent\textbf{Graph Neural Network (GNN):} Another current alternative that is applied is the temporal GNN. We used the work described in the paper of \cite{GNN}, and its hyperparameters. 
        
\section{Results}
\begin{figure}
             \centering
             \includegraphics[scale=0.55]{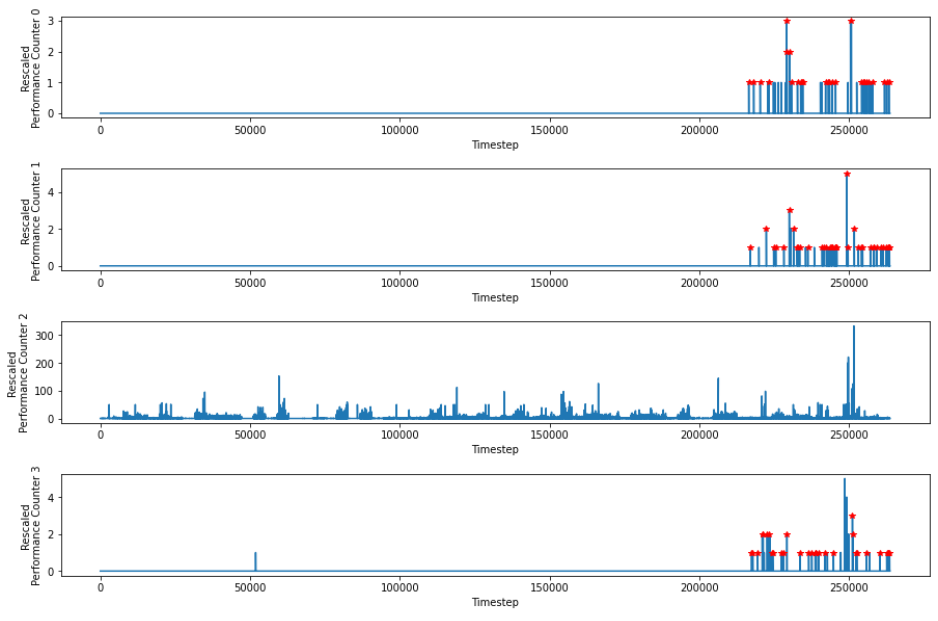}
             \caption{Reconstruction-based detection of anomalies by SSSDS4. The red marked stars are the samples where the anomaly correctly identified anomalies.}
             \label{fig:detection_results}
         \end{figure}

        Table \ref{tab:reconstruction_fullgeneve1_numresults} shows the numerical results of the anomaly detection.
        \begin{table}[!ht]
            \caption{The numerical results of the 
             anomaly detectors}
            \centering
            \begin{tabular}
            {|c|c|c|c|}
               \hline
               \textbf{Model} & \textbf{F1}  & \textbf{Precision} & \textbf{Recall} \\
               \hline
                SSSDS4 Reconstruction & $\mathbf{0.591}$ & $0.531$ & $0.666$ \\
                \hline
                SSSDS4 Forecasting & $0.265$ & $0.177$ & $0.523$ \\
               \hline
               Graph Neural Network & $0.540$ & $0.412$ & $\mathbf{0.535}$ \\
               \hline
               LSTM-AutoEncoder & $0.535$ & $0.405$ & $\mathbf{0.785}$ \\
               \hline
            \end{tabular}
            \label{tab:reconstruction_fullgeneve1_numresults}
            
        \end{table}
        The results show that the reconstruction anomaly detector achieves a 0.591 F1 score, outperforming the previously suggested GNN solution and the existing solution using LSTM-AutoEncoder regarding the F1 score. Interestingly, the GNN exhibits higher recall, so a complete solution might benefit from including both models. SSDS4 exhibits fewer false positives, while GNNs fewer false negatives. Figure \ref{fig:detection_results} illustrates some example true positives of the SSSDS4 Reconstruction model. Overall, SSSDS4 -Reconstruction exhibits the highest precision score of all present models, making it a great fit in situations where we prioritize minimizing type-I errors (false positives).
        
        SSSDS4 Forecasting does not exhibit the same level of performance, primarily due to its low precision score, indicating that the model struggles with false positives. A possible explanation is that given a sample, the difference between the ground-truth values and the predicted values by the model may be too high, which would exhibit greater values of \textit{Z-Score}, which in turn could produce a lot of false alarms assuming that the ground-truth values are not anomalies. This depends mainly on what the model has been conditioned on, the model assumes that the pattern in the data is mostly small values, which can produce difficulties in detecting anomalies since higher values are not necessarily anomalies.

        The model detects most of the anomalies shown in the plots (Figure \ref{fig:detection_results}); however, there are few setbacks from the anomaly detector. For instance, the model did not detect anomalies in the third-row plot, which exhibits a higher degree of value fluctuations compared to others. 
       In the last row of Figure \ref{fig:detection_results}, around timestep $50 000$, the model did not detect the sample as an anomaly even though the surrounding timesteps' values are significantly low. A possible explanation could be that the reconstruction values surrounding these timesteps resemble the anomaly's value; thus, the \textit{Z-Score}s around that timestep are small.

\begin{figure}
            \centering
            \includegraphics[scale=0.5]{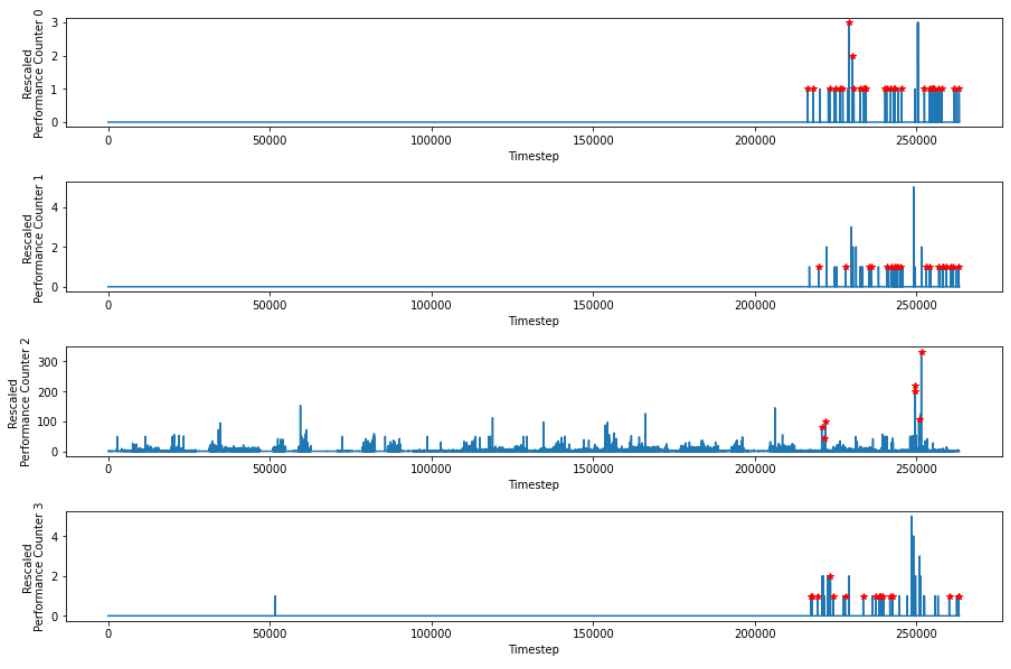}
            \caption{Forecasting-based detection of anomalies by SSSDS4. The red-marked stars are the samples where the anomaly correctly identified anomalies.}
            \label{fig:forecasting_detection_results}
            
{\addvspace{-1\baselineskip}}
        \end{figure}

        As for the forecasting-based anomaly detector, what is interesting to observe is in Figure \ref{fig:forecasting_detection_results} in the third-row plot.
        The forecasting-based anomaly detector managed to detect sample anomalies, whereas the reconstruction-based anomaly detector could not. This can be explained by how the forecasting-based anomaly detector uses conditional information to forecast the next vector-sequences. Conditional information can vary, thus, the model can adapt to this information and predict values that are aligned with the pattern of the data.

\section{Conclusion}    
    The paper presents a method to identify anomalies in telecom network multivariate time-series data, that combines time-series forecasting using diffusion models and post-processing techniques to find anomalies. The method achieved accurate and robust anomaly detection in operational telecom networks.
    Notably, the method presented in this paper pioneers the utilization of diffusion models in the context of Telecom Network anomaly detection. Importantly, our method has been assessed using operational network data. Our experiments on real-world datasets demonstrate that our approach outperforms state-of-the-art baselines performed on the same dataset in terms of F1 score.

    

    
    Given the models' performance and the problem context to which they have been applied, diffusion models show promise for improving software defect detection. By exploiting their ability to capture patterns and identify anomalies, these models can be effectively used to improve fault detection in software systems. 
    The paper demonstrates the suitability of diffusion models as an alternative for anomaly detection in the telecommunications domain. 
    To further improve the capabilities and applicability of diffusion models, future work should prioritize the following tasks: 
    
    - \textbf{Anomaly detection for data with different statistical properties}. Diffusion models in this work predominantly assume Gaussian distributions. However, future research should focus on developing techniques to handle non-Gaussian distributions to accommodate a wider range of data types, such as categorical data. TabDDPM \cite{tabddpm} added categorical future on general tabular (not time-series) datasets, yielding promising results. Applying it to time-series data could be a possibility that enhances the model's performance. This would enable diffusion models to deal effectively with time-series data with different statistical properties.

    - \textbf{Anomaly detection in multivariate settings and contextual anomaly management}. While diffusion models have shown promise in univariate time-series anomaly detection, extending their capabilities to multivariate settings is critical. Future work should develop methodologies enabling diffusion models to detect anomalies in complex, multivariate time-series data. In addition, the models should be improved to detect contextual anomalies that consider the anomaly's surrounding context to provide more accurate and meaningful results; this could be achieved by introducing the concept of latent diffusion model where the user can enter prompts that would allow the model to consider the textual information when detecting anomalies.
    
    - \textbf{Anomaly detection using Spatio-temporal Diffusion Models}. Telecom Networks are inherently distributed. Telecom Networks typically consist of multi-standard Radio nodes deployed in different geographic areas. Recently, diffusion models \cite{cachay2023dyffusion} showed promising results in forecasting tasks using spatio\-temporal data. Future work should evaluate methods to leverage the spatio\-temporal nature of Telecom Networks using diffusion models. 

\bibliographystyle{ieeetr}
\bibliography{reference}

\end{document}